\documentclass[runningheads]{llncs}
\usepackage[T1]{fontenc}
\usepackage{microtype} 
\usepackage{graphicx}
\usepackage[table]{xcolor}
\usepackage{colortbl}
\usepackage{hyperref}
\usepackage{booktabs}
\usepackage{multirow}
\usepackage{eqparbox}
\usepackage{arydshln}
\usepackage{amsmath}
\usepackage{cite}
\usepackage{tikz}
\usetikzlibrary{arrows, arrows.meta, calc, positioning, fit, backgrounds}
\usepackage{pgfplots}
\pgfplotsset{compat=1.18}
\definecolor{capa}{RGB}{180,50,30}
\definecolor{rfmain}{RGB}{210,140,30}       
\definecolor{rflight}{RGB}{255,243,218}
\definecolor{capamain}{RGB}{26,138,122}      
\definecolor{capalight}{RGB}{215,242,237}

\definecolor{rank1}{HTML}{1A9850}
\definecolor{rank2}{HTML}{66BD63}
\definecolor{rank3}{HTML}{A6D96A}
\definecolor{rank4}{HTML}{D4A028}
\definecolor{rank5}{HTML}{E06030}
\definecolor{rank6}{HTML}{B81D2E}
\definecolor{bg1}{HTML}{E5F5E0}
\definecolor{bg2}{HTML}{D5EBCD}
\definecolor{bg3}{HTML}{E8E8B0}
\definecolor{bg4}{HTML}{F0D898}
\definecolor{bg5}{HTML}{F0BB88}
\definecolor{bg6}{HTML}{E89B8B}
\definecolor{secbg}{HTML}{F5F5F5}
\newcommand{\auc}[2]{%
  \ifcase#2\or
    \cellcolor{bg1}#1\,{\scriptsize\textcolor{rank1}{\textbf{(#2)}}}\or
    \cellcolor{bg2}#1\,{\scriptsize\textcolor{rank2}{\textbf{(#2)}}}\or
    \cellcolor{bg3}#1\,{\scriptsize\textcolor{rank3}{\textbf{(#2)}}}\or
    \cellcolor{bg4}#1\,{\scriptsize\textcolor{rank4}{\textbf{(#2)}}}\or
    \cellcolor{bg5}#1\,{\scriptsize\textcolor{rank5}{\textbf{(#2)}}}\or
    \cellcolor{bg6}#1\,{\scriptsize\textcolor{rank6}{\textbf{(#2)}}}%
  \fi
}
\newcommand{\aucb}[2]{%
  \ifcase#2\or
    \cellcolor{bg1}\textbf{#1}\,{\scriptsize\textcolor{rank1}{\textbf{(#2)}}}\or
    \cellcolor{bg2}\textbf{#1}\,{\scriptsize\textcolor{rank2}{\textbf{(#2)}}}\or
    \cellcolor{bg3}\textbf{#1}\,{\scriptsize\textcolor{rank3}{\textbf{(#2)}}}\or
    \cellcolor{bg4}\textbf{#1}\,{\scriptsize\textcolor{rank4}{\textbf{(#2)}}}\or
    \cellcolor{bg5}\textbf{#1}\,{\scriptsize\textcolor{rank5}{\textbf{(#2)}}}\or
    \cellcolor{bg6}\textbf{#1}\,{\scriptsize\textcolor{rank6}{\textbf{(#2)}}}%
  \fi
}
\newcommand{\sch}[1]{%
  \ifcase#1\or
    \cellcolor{bg1}\textcolor{rank1}{\large\textbf{#1}}\or
    \cellcolor{bg2}\textcolor{rank2}{\large\textbf{#1}}\or
    \cellcolor{bg3}\textcolor{rank3}{\large\textbf{#1}}\or
    \cellcolor{bg4}\textcolor{rank4}{\large\textbf{#1}}\or
    \cellcolor{bg5}\textcolor{rank5}{\large\textbf{#1}}\or
    \cellcolor{bg6}\textcolor{rank6}{\large\textbf{#1}}%
  \fi
}
\newcommand{\schd}[1]{%
  \ifcase#1\or
    \cellcolor{bg1}\textcolor{rank1}{\large\textbf{#1}}$^{\!\dagger}$\or
    \cellcolor{bg2}\textcolor{rank2}{\large\textbf{#1}}$^{\!\dagger}$\or
    \cellcolor{bg3}\textcolor{rank3}{\large\textbf{#1}}$^{\!\dagger}$\or
    \cellcolor{bg4}\textcolor{rank4}{\large\textbf{#1}}$^{\!\dagger}$\or
    \cellcolor{bg5}\textcolor{rank5}{\large\textbf{#1}}$^{\!\dagger}$\or
    \cellcolor{bg6}\textcolor{rank6}{\large\textbf{#1}}$^{\!\dagger}$%
  \fi
}
\newcommand{\wtl}[2]{%
  \ifcase#1\or
    \cellcolor{bg1}#2\or
    \cellcolor{bg2}#2\or
    \cellcolor{bg3}#2\or
    \cellcolor{bg4}#2\or
    \cellcolor{bg5}#2\or
    \cellcolor{bg6}#2%
  \fi
}
\newcommand{\secrow}[1]{%
  \rowcolor{secbg}\multicolumn{8}{@{}l@{}}{\;\textsc{#1}}%
}
\begin{document}
\title{Active Learning for Cascaded Object Detection: Balancing Coverage and Uncertainty in Table Extraction Pipelines}
\titlerunning{Active Learning for Cascaded Object Detection}
%
\author{Eliott Thomas\inst{1,2}\orcidID{0009-0008-5266-8797} \and
Mickael Coustaty\inst{1}\orcidID{0000-0002-0123-439X} \and
Aurélie Joseph\inst{2}\orcidID{0000-0002-5499-6355} \and
Gaspar Deloin\inst{2}\orcidID{0009-0007-2449-5385} \and
Vincent Poulain d'Andecy\inst{2}\orcidID{0009-0008-5515-3561} \and
Jean-Marc Ogier\inst{1}\orcidID{0000-0002-5666-475X}}
\authorrunning{E. Thomas et al.}
\institute{L3i, La Rochelle Université, La Rochelle, France\\
\email{eliott.thomas@univ-lr.fr, mcoustat@univ-lr.fr, jean-marc.ogier@univ-lr.fr} \and
Yooz, Aimargues, France\\
\email{aurelie.joseph@getyooz.com, gaspar.deloin@getyooz.com, vincent.poulaindandecy@getyooz.com}}
\maketitle              

\begin{abstract}
  Table extraction from business documents relies on a cascaded pipeline where Table Detection (TD) first localizes tables and Table Structure Recognition (TSR) then recovers their internal layout. Building
  task-specific training sets for this pipeline is costly, particularly for TSR which requires fine-grained structural annotations. Active learning (AL) can reduce this annotation burden, yet most AL strategies
  are designed for single-model tasks and do not account for inter-stage dependencies in cascaded architectures. In this work, we present the first adaptation of Uncertainty Herding (UHerding), a hybrid
  coverage-uncertainty sampling method originally proposed for image classification, to cascaded object detection pipelines. We propose two pipeline-aware extensions that exploit the TD-to-TSR dependency:
  RankFusion adds dual-manifold coverage over both detection and structure representation spaces, while CAPA further incorporates stage-dependent gating and per-task uncertainty calibration.
  Extensive experiments across two public (PubTables-1M~\cite{CVPR_Smock_PubTables1M_2022} and FinTabNet~\cite{Zheng2020GlobalTE}) and two private table extraction datasets, with various annotation budgets (from 71 to 500 documents) show that UHerding generalizes well to table extraction, outperforming each baseline. Among pipeline-aware variants, RankFusion achieves higher expected gains but at the cost of greater variance, while CAPA emerges as the most consistent strategy,
  outperforming standard UHerding on three out of four datasets.

\end{abstract}

\section{Introduction}

  Business documents such as invoices, financial reports, and contracts rely on tables to organize structured data, making table extraction a central task in document analysis. The most common approach formulates
  this problem as a cascaded detection pipeline \cite{Kasem2022DeepLF,Schreiber2017DeepDeSRTDL,WACV_Eliott_RAPTOR_2025}: a Table Detection (TD) model first localizes tables within the document, and a Table
  Structure Recognition (TSR) model then recovers their internal layout, identifying rows, columns, and headers from the detected regions. Deploying such pipelines on new document domains requires task-specific
  training data, as publicly available table datasets \cite{CVPR_Smock_PubTables1M_2022,Zheng2020GlobalTE,Gao2019ICDAR2C,Gbel2013ICDAR2T} often differ significantly in layout, structure, and visual
  characteristics from the target domain. However, building these training sets is costly: while TD requires a single bounding box per table, TSR demands fine-grained annotations for each structural element,
  typically one header, several columns, and numerous rows per table. Active learning (AL) \cite{Settles2009ActiveLL,Gautam2025TableDW} offers a principled approach to this annotation burden: given a large pool
  of unlabeled documents, AL strategies select the most informative samples for annotation, maximizing model performance under a limited labeling budget.

  This cascaded formulation introduces a structural dependency: TSR operates on regions predicted by TD, so detection failures propagate irrecoverably downstream. The two stages also differ fundamentally in input scope, task complexity, and class structure: TD is single-class detection on full pages, while TSR is multi-class recognition on cropped table images. These asymmetries suggest that the two stages may benefit from different training signals and may not improve at the same rate as annotated data is added.

  Active learning has been widely studied as a means to reduce annotation costs in deep learning \cite{Settles2009ActiveLL,Ren2020ASO}. Existing strategies broadly fall into two families: coverage-based methods
  \cite{Sener2017ActiveLF,Yehuda2022ActiveLT,Bae2024GeneralizedCF}, which prioritize representing the diversity of the unlabeled pool, and uncertainty-based methods \cite{Lewis1994ASA,gal2016dropout,gal2017deep}, which favor samples on which the model is least confident. These two paradigms tend to excel in different budget regimes, with coverage dominating in low-budget settings and uncertainty becoming
  more effective as the labeled set grows \cite{hacohen2022active}. Uncertainty Herding (UHerding) \cite{bae2025uncertainty} was recently proposed to bridge this gap through an adaptive mechanism that balances
  coverage and uncertainty based on the current training state, making it a strong baseline across budget regimes. However, UHerding was designed for image classification with a single model. To the best of our
  knowledge, this work presents the first adaptation of UHerding to a cascaded object detection pipeline, where two interdependent models must be trained jointly from a shared pool of annotated documents.

  While adapted UHerding performs well in this setting, it remains pipeline-agnostic: it selects samples based on a single shared representation and a combined uncertainty signal, without distinguishing between
  the contributions of each stage. Given the structural asymmetry between TD and TSR, a natural question arises: can AL strategies that explicitly account for inter-stage dependencies provide consistent
  improvements, and what trade-offs emerge? Drawing on prior work on active learning for pipeline models \cite{roth2008active}, we propose two pipeline-aware extensions. RankFusion introduces dual-manifold
  coverage, selecting samples that are informative in both the detection and structure representation spaces. CAPA goes further by incorporating stage-dependent gating and per-task uncertainty
  calibration, dynamically adapting to the reliability of each pipeline stage. Our contributions are as follows:

  \begin{enumerate}
      \item To the best of our knowledge, the first adaptation of UHerding to a cascaded object detection pipeline for table extraction.
      \item Two pipeline-aware extensions of our adaptation of UHerding: RankFusion, adding dual-manifold coverage over document-level and table-level spaces, and CAPA, extending RankFusion with stage-dependent gating and per-task uncertainty calibration.
      \item An extensive evaluation across four diverse table extraction datasets, two public (PubTables-1M~\cite{CVPR_Smock_PubTables1M_2022} and FinTabNet~\cite{Zheng2020GlobalTE}) and two private, with 10 seeds per configuration, revealing a consistency versus expected gain trade-off among
  pipeline-aware strategies.
  \end{enumerate}

\section{Related Work}

  \subsection{Table Extraction}

    Table extraction has been extensively studied \cite{Kasem2022DeepLF}, with approaches broadly categorized into detection-based cascaded pipelines
  \cite{Schreiber2017DeepDeSRTDL,WACV_Eliott_RAPTOR_2025,ESA_Xiao_CascadeTSRDet_2025} and generative methods based on vision-language models \cite{Nassar_2025_ICCV,blecher2023nougat,kim2022donut} that
  directly output structured markup such as HTML or LaTeX. While generative approaches have shown strong results on academic documents, cascaded pipelines remain the dominant paradigm for applications requiring
  spatial coordinates of individual structural elements. Within this paradigm, TD and TSR are typically addressed as separate object detection tasks with distinct characteristics.

  For table detection, most recent approaches rely on general-purpose object detectors adapted for document images, such as DETR \cite{ECCV_Carion_DETR_2020}, D-FINE \cite{ICLR_Peng_DFINE_2025}, and YOLOv9
  \cite{wang2024yolov9}. Smock et al. \cite{CVPR_Smock_PubTables1M_2022} showed that such architectures can be effectively trained for both TD and TSR on large-scale table datasets. For structure recognition, the
  geometric regularity of table layouts, with repeating rows, columns, and headers, has additionally motivated a range of task-specific architectures \cite{PR_LONG_LOREPlus_2025,IJCAI_KhangTFLOP_2024,EMNLP_Zhang_UniTabNet_2024,ICDAR_Hou_TABLET_2026}.

Standard benchmarks \cite{Gbel2013ICDAR2T,Gao2019ICDAR2C} remain small, while PubTables-1M \cite{CVPR_Smock_PubTables1M_2022} and FinTabNet \cite{Zheng2020GlobalTE} better support AL research. GriTS \cite{ICDAR_Smock_GriTS_2023} addresses inconsistencies in the earlier TEDS metric \cite{zhong2020image}. Prior work reduces labeling cost through semi-supervised learning for layout analysis \cite{banerjee2024semidocseg}, table detection \cite{shehzadi2024towards}, and full pipelines \cite{ICDAR_ThomasQUEST_2025}. To our knowledge, Gautam et al. \cite{Gautam2025TableDW} provide the only AL application to table extraction, limited to detection. No prior work addresses the full cascaded pipeline.

  \subsection{Active Learning for Deep Learning}

  Uncertainty sampling \cite{Lewis1994ASA} selects samples on which the current model is most uncertain, through measures such as prediction confidence, margin, or entropy \cite{Settles2009ActiveLL}. For deep networks, Bayesian approximations such as Monte Carlo Dropout \cite{gal2016dropout,gal2017deep} enable uncertainty estimation without architectural changes.

  Coverage-based methods take the opposite approach, prioritizing samples that represent the diversity of the unlabeled pool. CoreSet \cite{Sener2017ActiveLF} formulates this as a k-center problem in a learned feature space. ProbCover \cite{Yehuda2022ActiveLT} refines this through fixed-radius coverage in embedding space. MaxHerding \cite{Bae2024GeneralizedCF} further improves upon this line through kernel herding with a Gaussian kernel, providing smoother coverage estimates more robust to hyperparameter choices.

This complementarity between uncertainty and coverage methods across budget regimes, formalized by Hacohen et al. \cite{hacohen2022active}, has motivated hybrid approaches that combine both signals. BADGE
  \cite{ash2020badge} captures both signals jointly through gradient embeddings, though its reliance on task-specific gradients makes it difficult to extend to multi-model pipelines. UHerding \cite{bae2025uncertainty} instead explicitly combines coverage with uncertainty weighting and automatically adapts
  the balance based on the current training state, removing the need for manual regime identification.

Active learning has been applied to document layout analysis \cite{subramanian2023tactful,shen2022olala}, but its application to table extraction remains limited. Recent work has also addressed the characterization of AL performance curves. PALM \cite{machnio2025label} fits a parametric model to learning trajectories for cross-strategy comparison, though reliable fits require learning curves that extend into the saturation regime, limiting applicability to settings with broad budget ranges.

  \subsection{Active Learning for Multi-Stage Models}

  While active learning has been extensively studied for individual models, its application to multi-stage pipelines has received comparatively little attention. Roth and Small \cite{roth2008active} proposed a
  beta-weighting framework for active learning in pipeline models, where the contribution of each stage to the overall loss is weighted by the reliability of upstream components. A key insight from their work is
  that early pipeline stages should be prioritized until they perform reliably, after which the learning focus should shift to later stages. Despite the theoretical appeal of this framework, it has seen limited
  follow-up, and no prior work has applied pipeline-aware active learning to object detection architectures. This gap motivates the pipeline-aware extensions we introduce in the following section.

  \section{Method}

  \subsection{Problem Formulation}
  \label{sec:problem_formulation}

  We consider a table extraction pipeline composed of two models trained from a shared set of annotated documents. A Table Detection model $M_{\text{td}}$ localizes tables within full document pages, and a Table Structure Recognition model $M_{\text{tsr}}$ identifies rows, columns, and headers within each detected region. These two stages are coupled at inference: $M_{\text{tsr}}$ operates on the table regions predicted by $M_{\text{td}}$, so the end-to-end performance of the pipeline is bounded by the quality of the upstream detection. We scope the pipeline to single-table documents without spanning cells, yielding a regular grid structure and three structural element classes for TSR: rows, columns, and headers. This design isolates the effect of the sampling strategy from confounding factors such as multi-table disambiguation or complex cell merging. As our results show, even this well-defined setting proves challenging at the annotation budgets considered, with models far from saturation across all datasets.

  We formulate active learning for this pipeline in the standard pool-based setting. Let $\mathcal{U} = \{x_n\}_{n=1}^{N}$ denote a large pool of unlabeled documents and $\mathcal{L}_0$ a small initial labeled set. At each iteration $t$, both models are trained on $\mathcal{L}_t$: $M_{\text{td}}$ is trained on full document pages with table bounding box annotations, while $M_{\text{tsr}}$ is trained on ground-truth table crops with structural element labels, following the standard setup in table structure recognition where structural annotations are provided at the table level. A sampling strategy then selects a batch $\mathcal{S}_t \subset \mathcal{U}_t$ of documents for annotation, and the sets are updated as $\mathcal{L}_{t+1} = \mathcal{L}_t \cup \mathcal{S}_t$ and $\mathcal{U}_{t+1} = \mathcal{U}_t \setminus \mathcal{S}_t$. This process repeats for $T$ iterations with a fixed batch size $|\mathcal{S}_t| = b$, yielding a final labeled set $\mathcal{L}_T$ of size $|\mathcal{L}_0| + Tb$.

  Annotating a single document provides training signal for both stages, but the two stages do not necessarily benefit equally from the same document. Moreover, the relative contribution of each stage to the overall pipeline error is dataset-dependent: in some domains $M_{\text{td}}$ is the bottleneck, while in others $M_{\text{tsr}}$ limits end-to-end performance.

  The objective is to select, at each iteration, the batch $\mathcal{S}_t$ that maximizes end-to-end pipeline performance, evaluated by a metric that measures structural correctness of extracted tables through grid-based cell matching. This requires the sampling strategy to balance informativeness across both stages rather than optimizing for a single model.

  \subsection{Coverage-Uncertainty Sampling for Cascaded Pipelines}
  \label{sec:uherding}

  Recent work in single-model active learning has shown that combining coverage and uncertainty signals outperforms either signal alone across budget regimes~\cite{bae2025uncertainty}. We build on this principle and adapt it to the cascaded detection setting, which requires extending both the coverage and uncertainty components to account for multiple stages with different embedding spaces and error profiles.

  Generalized coverage (GCoverage)~\cite{Bae2024GeneralizedCF} measures how well a labeled set $\mathcal{L}$ represents the unlabeled pool. Let $g$ be a pretrained document encoder that maps each document to a $d$-dimensional representation, fixed throughout the AL process and independent of the detection models $M_{\text{td}}$ and $M_{\text{tsr}}$. Given a Gaussian kernel $k_\sigma(x, x'; g) = \exp\bigl({-\|g(x) - g(x')\|^2}/{\sigma^2}\bigr)$, where $\|\cdot\|$ denotes the Euclidean norm, the estimated GCoverage of $\mathcal{L}$ over the pool $\mathcal{U}$ is:
  \begin{equation}
    \widehat{C}_{k_\sigma}(\mathcal{L}) = \frac{1}{N} \sum_{n=1}^{N}
    \max_{x' \in \mathcal{L}} k_\sigma(x_n, x'; g).
    \label{eq:gcoverage}
  \end{equation}
  MaxHerding~\cite{Bae2024GeneralizedCF} greedily adds the candidate that most increases $\widehat{C}_{k_\sigma}$.

  UHerding~\cite{bae2025uncertainty} extends this objective by weighting coverage gains with a nonnegative uncertainty function $U$, so that uncertain regions of the pool receive higher priority. For each pool document $x_n$, let $h_n = \max_{x' \in \mathcal{L}_t} k_\sigma(x_n, x'; g)$ denote its current best coverage from the labeled set. The coverage gain from a candidate $\tilde{x}$ at pool point $x_n$ is $\max(0,\; k_\sigma(x_n, \tilde{x}; g) - h_n)$: positive only when $\tilde{x}$ is closer to $x_n$ in embedding space than any currently labeled document. The next point is selected as:
  \begin{equation}
    x^* = \arg\max_{\tilde{x} \in \mathcal{U}_t} \sum_{n=1}^{N}
    U(x_n) \cdot \max\bigl(0,\; k_\sigma(x_n, \tilde{x}; g) - h_n\bigr).
    \label{eq:uherding}
  \end{equation}
  After selecting $x^*$, the coverage values are updated as $h_n \leftarrow \max(h_n, k_\sigma(x_n, x^*; g))$ for all $n$, and the process repeats until $\mathcal{S}_t$ is filled. This procedure selects documents that best cover uncertain, underrepresented regions of the pool.

  We adapt UHerding to cascaded table extraction as follows. For the embedding function $g$, the original work~\cite{bae2025uncertainty} uses self-supervised image features (e.g., SimCLR~\cite{ICML_Chen_SimCLR_2020}, DINO~\cite{ICCV_Caron_DINO_2021}) designed for natural images. We replace these with a pretrained document image encoder that captures layout and visual properties specific to document pages, while remaining independent of any table-related task. This choice of $g$ is instantiated in Section~\ref{sec:experimental_setup}.

  For uncertainty, Bae et al.~\cite{bae2025uncertainty} showed that the choice of measure (margin, entropy, confidence) has limited impact in the single-model setting. We adopt confidence-based uncertainty, extended to both pipeline stages. Let $c_{\text{top}}(x)$ be the confidence of the highest-scoring detection in document $x$, and $\bar{c}_{\text{tsr}}(x)$ the mean confidence across retained structural predictions. When no prediction exceeds a threshold $t$, uncertainty defaults to~1:
  \begin{equation}
    U_{\text{td}}(x) = \begin{cases} 1 - c_{\text{top}}(x) & c_{\text{top}} \!\geq\! t \\ 1 & \text{else} \end{cases} \quad
    U_{\text{tsr}}(x) = \begin{cases} 1 - \bar{c}_{\text{tsr}}(x) & \bar{c}_{\text{tsr}} \!\geq\! t \\ 1 & \text{else} \end{cases}
    \label{eq:stage_uncertainty}
  \end{equation}
  The combined uncertainty is:
  \begin{equation}
    U(x) = \frac{1}{2}\bigl(U_{\text{td}}(x) + U_{\text{tsr}}(x)\bigr),
    \label{eq:combined_uncertainty}
  \end{equation}
  treating both stages as equally informative. We relax this assumption in the pipeline-aware variants of Section~\ref{sec:pipeline_aware}.

  Two adaptive parameters control the balance between coverage and uncertainty at each AL iteration. The kernel bandwidth $\sigma_t$ determines the spatial scale of coverage in the embedding space of $g$: as the labeled set grows, $\sigma_t$ decreases, making coverage increasingly local and shifting the selection signal toward uncertainty. The original UHerding~\cite{bae2025uncertainty} sets $\sigma_t$ to the minimum pairwise distance in the labeled set:
  \begin{equation}
    \sigma_t^{\text{orig}} = \min_{\substack{u, v \in \mathcal{L}_t \\ u \neq v}}
    \|g(u) - g(v)\|.
    \label{eq:sigma_orig}
  \end{equation}
  Near-duplicate documents can make this minimum arbitrarily small, collapsing the kernel prematurely. We instead use the 5th percentile of pairwise distances:
  \begin{equation}
    \sigma_t = P_5\bigl(\{\|g(u) - g(v)\| : u, v \in \mathcal{L}_t,\; u \neq v\}\bigr),
    \label{eq:adaptive_sigma}
  \end{equation}
  which retains the decreasing behavior while being robust to outlier pairs.

  The calibration parameter $\tau_t$ controls how strongly uncertainty influences selection. The original UHerding~\cite{bae2025uncertainty} calibrates $\tau$ by minimizing Expected Calibration Error (ECE)~\cite{ICML_Guo_Calibration_2017} on a classifier's logits, which assumes a single model producing one probability distribution per input. In object detection, each document yields multiple predictions with individual confidence scores, and the per-document uncertainties $U_{\text{td}}$, $U_{\text{tsr}}$ are aggregations over these predictions, for which ECE has no standard formulation. We therefore adopt a rank-correlation approach: at each iteration, we compute the Spearman correlation $\rho_t$ between $U(x)$ and the per-document prediction error $1 - F_1(x)$, where $F_1$ is the GriTS-Con F1, on the validation subset of $\mathcal{L}_t$, and set:
  \begin{equation}
    \tau_t = \tau_{\min} + \bigl(1 - \max(0, \rho_t)\bigr) \cdot
    (\tau_{\max} - \tau_{\min}).
    \label{eq:adaptive_tau}
  \end{equation}
  Calibrated uncertainties are obtained as $U_{\text{cal}}(x) = U(x)^{1/\tau_t}$. When uncertainty correlates well with actual error ($\rho_t$ high), $\tau_t$ is small and the raw uncertainty signal is preserved. When the correlation is weak, $\tau_t$ is large, compressing the distribution toward uniform values and recovering coverage-dominated selection.

  \subsection{Exploiting the Pipeline Dependency}
  \label{sec:pipeline_aware}

  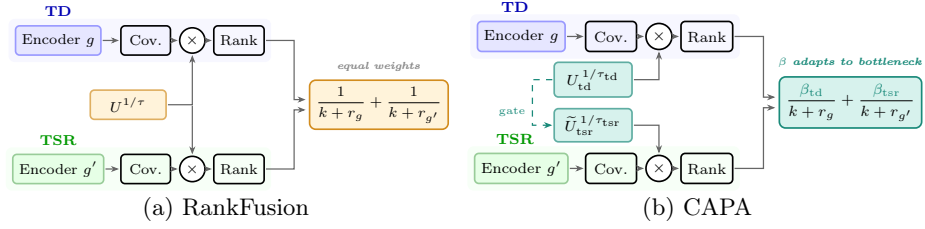
\begin{figure}[t]
    \centering
    \begin{minipage}[b]{0.48\linewidth}
    \centering
    \resizebox{\linewidth}{!}{%
    \begin{tikzpicture}[
      >=Stealth,
      box/.style={draw, thick, rounded corners=2pt, minimum height=0.5cm,
                  minimum width=0.85cm, align=center, font=\scriptsize},
      op/.style={circle, draw, thick, fill=white, inner sep=1.5pt,
                 font=\scriptsize\bfseries},
      arr/.style={-{Stealth[length=4pt,width=3pt]}, semithick, black!60,
                  shorten >=0.5pt, shorten <=0.5pt},
      stlbl/.style={font=\scriptsize\bfseries, text=black!55},
      note/.style={font=\tiny, text=black!45},
    ]
    \node[box, fill=blue!10, draw=blue!40, minimum width=1.1cm]
      (enc_td) at (0, 0) {Encoder $g$};
    \node[stlbl, text=blue!70!black, above=1pt of enc_td] {TD};
    \node[box, fill=blue!5] (cov_td) at (1.5, 0) {Cov.};
    \node[op] (mul_td) at (2.3, 0) {$\times$};
    \node[box] (rank_td) at (3.1, 0) {Rank};
    \node[box, fill=green!10, draw=green!50!black!40, minimum width=1.1cm]
      (enc_tsr) at (0, -2.2) {Encoder $g'$};
    \node[stlbl, text=green!60!black, above=1pt of enc_tsr] {TSR};
    \node[box, fill=green!5] (cov_tsr) at (1.5, -2.2) {Cov.};
    \node[op] (mul_tsr) at (2.3, -2.2) {$\times$};
    \node[box] (rank_tsr) at (3.1, -2.2) {Rank};
    \node[box, fill=rflight, draw=rfmain!70, minimum width=1.3cm]
      (unc) at (1.2, -1.1) {$U^{1/\tau}$};
    \node[draw=rfmain, thick, rounded corners=3pt, fill=rflight,
          minimum height=0.7cm, minimum width=1.6cm,
          align=center, font=\scriptsize]
      (fuse) at (5.5, -1.1)
      {$\dfrac{1}{k+r_g} + \dfrac{1}{k+r_{g'}}$};
    \draw[arr] (enc_td) -- (cov_td);
    \draw[arr] (enc_tsr) -- (cov_tsr);
    \draw[arr] (cov_td) -- (mul_td);
    \draw[arr] (cov_tsr) -- (mul_tsr);
    \draw[arr] (mul_td) -- (rank_td);
    \draw[arr] (mul_tsr) -- (rank_tsr);
    \draw[arr] (unc.east) -- (2.3, -1.1) -- (mul_td.south);
    \draw[arr] (2.3, -1.1) -- (mul_tsr.north);
    \draw[arr, shorten >=1.5pt] (rank_td.east) -- ++(0.5, 0) |- ([yshift=2.5pt]fuse.west);
    \draw[arr, shorten >=1.5pt] (rank_tsr.east) -- ++(0.5, 0) |- ([yshift=-2.5pt]fuse.west);
    \node[note, above=1pt of fuse, font=\tiny\bfseries] {\textit{equal weights}};
    \begin{scope}[on background layer]
      \node[fit=(enc_td)(rank_td), fill=blue!3, rounded corners=3pt, inner sep=3pt] {};
      \node[fit=(enc_tsr)(rank_tsr), fill=green!3, rounded corners=3pt, inner sep=3pt] {};
    \end{scope}
    \end{tikzpicture}%
    }%
    \\[-2pt]
    {\small (a) RankFusion}
    \end{minipage}\hfill
    \begin{minipage}[b]{0.50\linewidth}
    \centering
    \resizebox{\linewidth}{!}{%
    \begin{tikzpicture}[
      >=Stealth,
      box/.style={draw, thick, rounded corners=2pt, minimum height=0.5cm,
                  minimum width=0.85cm, align=center, font=\scriptsize},
      op/.style={circle, draw, thick, fill=white, inner sep=1.5pt,
                 font=\scriptsize\bfseries},
      arr/.style={-{Stealth[length=4pt,width=3pt]}, semithick, black!60,
                  shorten >=0.5pt, shorten <=0.5pt},
      garr/.style={-{Stealth[length=4pt,width=3pt]}, semithick, dashed, capamain,
                   shorten >=0.5pt, shorten <=0.5pt},
      stlbl/.style={font=\scriptsize\bfseries, text=black!55},
      note/.style={font=\tiny},
    ]
    \node[box, fill=blue!10, draw=blue!40, minimum width=1.1cm]
      (enc_td) at (0, 0) {Encoder $g$};
    \node[stlbl, text=blue!70!black, above=1pt of enc_td, xshift=-3pt] {TD};
    \node[box, fill=blue!5] (cov_td) at (1.5, 0) {Cov.};
    \node[op] (mul_td) at (2.3, 0) {$\times$};
    \node[box] (rank_td) at (3.1, 0) {Rank};
    \node[box, fill=green!10, draw=green!50!black!40, minimum width=1.1cm]
      (enc_tsr) at (0, -2.2) {Encoder $g'$};
    \node[stlbl, text=green!60!black, above=1pt of enc_tsr, xshift=-3pt] {TSR};
    \node[box, fill=green!5] (cov_tsr) at (1.5, -2.2) {Cov.};
    \node[op] (mul_tsr) at (2.3, -2.2) {$\times$};
    \node[box] (rank_tsr) at (3.1, -2.2) {Rank};
    \node[box, fill=capalight, draw=capamain!70, minimum width=1.3cm]
      (unc_td) at (1.2, -0.72) {$U_\text{td}^{\,1/\tau_\text{td}}$};
    \node[box, fill=capalight, draw=capamain!70, minimum width=1.3cm]
      (unc_tsr) at (1.2, -1.48) {$\widetilde{U}_\text{tsr}^{\,1/\tau_\text{tsr}}$};
    \draw[garr] (unc_td.west) -- ++(-0.35, 0) |- (unc_tsr.west)
      node[pos=0.35, left, font=\tiny, text=capamain] {gate};
    \node[draw=capamain, thick, rounded corners=3pt, fill=capalight,
          minimum height=0.7cm, minimum width=1.8cm,
          align=center, font=\scriptsize]
      (fuse) at (5.5, -1.1)
      {$\dfrac{{\color{capamain}\beta_\text{td}}}{k+r_g} + \dfrac{{\color{capamain}\beta_\text{tsr}}}{k+r_{g'}}$};
    \draw[arr] (enc_td) -- (cov_td);
    \draw[arr] (enc_tsr) -- (cov_tsr);
    \draw[arr] (cov_td) -- (mul_td);
    \draw[arr] (cov_tsr) -- (mul_tsr);
    \draw[arr] (mul_td) -- (rank_td);
    \draw[arr] (mul_tsr) -- (rank_tsr);
    \draw[arr] (unc_td.east) -- (2.3, -0.72) -- (mul_td.south);
    \draw[arr] (unc_tsr.east) -- (2.3, -1.48) -- (mul_tsr.north);
    \draw[arr, shorten >=1.5pt] (rank_td.east) -- ++(0.5, 0) |- ([yshift=2.5pt]fuse.west);
    \draw[arr, shorten >=1.5pt] (rank_tsr.east) -- ++(0.5, 0) |- ([yshift=-2.5pt]fuse.west);
    \node[note, text=capamain, above=1pt of fuse, font=\tiny\bfseries] {\textit{$\beta$ adapts to bottleneck}};
    \begin{scope}[on background layer]
      \node[fit=(enc_td)(rank_td), fill=blue!3, rounded corners=3pt, inner sep=3pt] {};
      \node[fit=(enc_tsr)(rank_tsr), fill=green!3, rounded corners=3pt, inner sep=3pt] {};
    \end{scope}
    \end{tikzpicture}%
    }%
    \\[-2pt]
    {\small (b) CAPA}
    \end{minipage}
    \caption{Overview of the two pipeline-aware selection strategies.
    Both score documents along a \textcolor{blue!70!black}{\textbf{Table Detection}}
    (\textcolor{blue!70!black}{TD}, blue) and a
    \textcolor{green!60!black}{\textbf{Table Structure Recognition}}
    (\textcolor{green!60!black}{TSR}, green) path, each computing
    coverage gains weighted by calibrated uncertainty~($\times$),
    then ranking and fusing via Reciprocal Rank Fusion.
    \textbf{(a)}~\textcolor{rfmain}{RankFusion} uses shared calibration~$\tau$ and
    equal fusion weights.
    \textbf{(b)}~\textcolor{capamain}{CAPA} adds per-stage calibration
    ($\tau_\text{td}$,~$\tau_\text{tsr}$), gates
    \textcolor{green!60!black}{TSR} uncertainty by
    \textcolor{blue!70!black}{TD} confidence
    (\textcolor{capamain}{dashed}), and adapts fusion
    weights~$\beta$ to the current bottleneck.}
    \label{fig:capa_overview}
  \end{figure}

  The formulation of Section~\ref{sec:uherding} treats both pipeline stages symmetrically: coverage operates in a single embedding space, uncertainties are averaged, and a single calibration parameter governs the trade-off. However, in a cascaded pipeline, this symmetry does not hold. The two stages can exhibit markedly different difficulty levels, and they are not independent: TSR can only succeed on documents where TD correctly detects a table. When table detection is the bottleneck, investing in TSR diversity yields diminishing returns because most tables are missed before structure recognition begins. Conversely, when detection is reliable, structure recognition errors dominate. We introduce two extensions that progressively relax these symmetry assumptions. Figure~\ref{fig:capa_overview} provides an overview.

  \subsubsection{RankFusion.}

  A first limitation of single-manifold coverage is that the document-level encoder $g$ captures global page layout but is agnostic to table-internal structure. We introduce a second encoder $g'$ that operates at the table level, producing representations sensitive to structural elements (rows, columns, headers). Coverage is now computed independently in both spaces, each with its own adaptive bandwidth $\sigma_t$ and $\sigma'_t$ (Eq.~\ref{eq:adaptive_sigma}). For a candidate $\tilde{x}$, let $I_g(\tilde{x})$ and $I_{g'}(\tilde{x})$ denote the uncertainty-weighted coverage gains in the document-level and table-level spaces, respectively. Rather than summing these heterogeneous scores, which would require careful normalization across spaces of different dimensionality and scale, we convert each to a rank among the remaining candidates and combine them via Reciprocal Rank Fusion (RRF)~\cite{SIGIR_Cormack_RRF_2009}:
  \begin{equation}
    \text{score}(\tilde{x}) = \frac{1}{k + r_g(\tilde{x})} + \frac{1}{k + r_{g'}(\tilde{x})},
    \label{eq:rrf}
  \end{equation}
  where $r_g$ and $r_{g'}$ are the ranks in each space and $k$ is a smoothing constant. This rank-based fusion is scale-invariant and does not require the two embedding spaces to be commensurable. We call this variant \textbf{RankFusion}. It retains UHerding's combined uncertainty $U$ (Eq.~\ref{eq:combined_uncertainty}) and single calibration parameter $\tau_t$, extending only the coverage component to dual manifolds.

  \subsubsection{CAPA.}

  RankFusion weights both spaces equally, which does not account for the pipeline dependency: TSR can only succeed on documents where TD correctly detects a table. Following Roth and Small~\cite{roth2008active}, who showed that early-stage errors propagate through pipeline models and that active learning should prioritize the bottleneck stage, we introduce \textbf{CAPA} (Calibrated and Pipeline-Aware), which extends RankFusion along three axes. First, a global weighting mechanism directs sampling effort toward the current bottleneck. Let $\text{AP}_{\text{td}}$ and $\text{AP}_{\text{tsr}}$ denote the validation performance of each stage at iteration $t$. We define:
  \begin{equation}
    \beta_{\text{td}} \propto 1 - \text{AP}_{\text{td}}, \quad
    \beta_{\text{tsr}} \propto (1 - \text{AP}_{\text{tsr}}) \cdot \text{AP}_{\text{td}},
    \label{eq:beta_weights}
  \end{equation}
  normalized such that $\beta_{\text{td}} + \beta_{\text{tsr}} = 1$. The $\text{AP}_{\text{td}}$ factor in $\beta_{\text{tsr}}$ encodes the pipeline dependency: when detection is poor, improving structure recognition has limited value. These weights are applied to the RRF fusion as $\beta_{\text{td}} / (k + r_g) + \beta_{\text{tsr}} / (k + r_{g'})$.

  Second, CAPA replaces the combined uncertainty $U$ (Eq.~\ref{eq:combined_uncertainty}) with stage-specific signals. For the table detection branch, $U_{\text{td}}$ is used directly. For the structure recognition branch, a local gating mechanism suppresses TSR uncertainty for documents where detection is itself unreliable:
  \begin{equation}
    \widetilde{U}_{\text{tsr}}(x) = \bigl(1 - U_{\text{td}}(x)\bigr) \cdot U_{\text{tsr}}(x).
    \label{eq:gating}
  \end{equation}
  When $U_{\text{td}}$ is high, the factor $(1 - U_{\text{td}})$ approaches zero, suppressing the TSR signal regardless of the raw TSR uncertainty and preventing wasted structural annotations on documents with missed detections.

  Third, CAPA replaces the single calibration parameter $\tau_t$ with stage-specific values $\tau_{\text{td}}$ and $\tau_{\text{tsr}}$, each computed via Spearman correlation between the corresponding uncertainty signal and the stage-specific prediction error on the validation subset of $\mathcal{L}_t$ (Eq.~\ref{eq:adaptive_tau}). Crucially, gating is applied before calibration: if the order were reversed, a large $\tau_{\text{tsr}}$ would first compress $U_{\text{tsr}}$ toward uniform values, and the subsequent gate would merely scale a near-constant quantity by detection confidence rather than modulate the actual uncertainty signal.

  \section{Experiments}

  \subsection{Datasets}
  \label{sec:datasets}

  We evaluate on four table extraction datasets spanning diverse document types (Table~\ref{tab:datasets}). Two are publicly available: \textbf{PubTables-1M}~\cite{CVPR_Smock_PubTables1M_2022}, a large-scale dataset of academic papers, and \textbf{FinTabNet}~\cite{Zheng2020GlobalTE}, containing financial reports. Two are proprietary: \textbf{invoice-int}, consisting of invoices, and \textbf{business-int}, a collection of mixed business documents. The two proprietary datasets cannot be released, so they serve only to test robustness on real-world business documents. Every methodological claim is also validated on the two public benchmarks, and the full pipeline relies on publicly available models (D-FINE for detection, DiT for embeddings), so the core results remain independently reproducible. Following the scope defined in Section~\ref{sec:problem_formulation}, we retain only documents containing exactly one table with no spanning cells. For PubTables-1M, we apply this filter to the original validation split, and for FinTabNet to the training split, in both cases obtaining pools large enough for meaningful AL experiments while keeping the per-iteration scoring of the full unlabeled pool computationally tractable.

  \begin{table}[t]
    \centering
    \caption{Dataset overview. \emph{Pool} denotes the unlabeled documents available for AL selection. \emph{Test} is a held-out evaluation set.}
    \label{tab:datasets}
    \setlength{\tabcolsep}{8pt}
    \begin{tabular}{llrrrc}
      \toprule
      Dataset & Domain & Total & Pool & Test & Public \\
      \midrule
      PubTables-1M & Academic & 4{,}459 & 3{,}459 & 1{,}000 & Yes \\
      FinTabNet & Financial & 8{,}066 & 7{,}105 & 961 & Yes \\
      \hdashline
      \noalign{\vskip 2pt}
      invoice-int & Invoices & 2{,}394 & 2{,}117 & 277 & No \\
      business-int & Business & 3{,}920 & 2{,}920 & 1{,}000 & No \\
      \bottomrule
    \end{tabular}
  \end{table}

  \subsection{Experimental Setup}
  \label{sec:experimental_setup}

  Both pipeline stages use D-FINE-X~\cite{ICLR_Peng_DFINE_2025}, a DETR-based detector initialized from a COCO-pretrained checkpoint and trained separately for table detection (1 class) and table structure recognition (3 classes: row, column, header). This training is the active learning loop itself, not a separate pre-training stage: at each round both models are re-initialized from the COCO checkpoint and trained on the current labeled set, with no warm-starting from the previous round. Models are trained with AdamW (learning rate $2.5 \times 10^{-4}$, weight decay $1.25 \times 10^{-4}$), batch size 24, and standard geometric and photometric augmentations (zoom-out, IoU crop, brightness and contrast jitter). Training runs for up to 100 epochs with early stopping on validation loss (patience 40) to avoid retaining a local optimum. The document-level encoder $g$ (Section~\ref{sec:uherding}) is DiT~\cite{ACMMM_Li_DiT_2022}, a Vision Transformer pretrained on document images, providing fixed representations throughout the AL process. The table-level encoder $g'$ (Section~\ref{sec:pipeline_aware}) reuses the multi-scale features of the trained TSR model, which are globally average-pooled and concatenated into a single 1152-dimensional vector. This pairing was selected after preliminary analysis indicated that combining document-level and structure-level representations yielded the best coverage balance. Unlike $g$, these representations evolve as the model is retrained at each iteration. Each experiment begins with 71 randomly sampled documents and grows the labeled set over six sampling rounds, yielding seven equally-spaced budget levels up to 500 documents, balancing evaluation granularity with computational cost. At each round, 10\% of the labeled set is held out for validation and for computing adaptive parameters ($\sigma_t$, $\tau_t$, $\beta$). The RRF smoothing constant is set to $k = 60$~\cite{SIGIR_Cormack_RRF_2009}, the calibration bounds to $\tau_{\min} = 1$, $\tau_{\max} = 20$, and predictions with confidence below $t = 0.3$ are discarded before computing uncertainties. Unlike image classification AL with fixed training schedules~\cite{hacohen2022active,Bae2024GeneralizedCF,ash2020badge}, transformer-based detection requires a validation holdout for early stopping. As L\"uth et al.~\cite{NeurIPS_Luth_Pitfalls_2023} note, validation protocols are rarely reported yet can matter as much as multiplying the labeled budget fivefold. All configurations share the same initial set per seed, and each is repeated over 10 seeds (independent initial labeled sets) to enable paired comparisons.

  End-to-end performance is measured by GriTS-Con F1~\cite{ICDAR_Smock_GriTS_2023}, which evaluates both structural correctness and text content of extracted cells. Run-to-run variance is high at low budgets across all strategies. We therefore summarize each strategy by the normalized area under its median learning curve rather than the mean, an average F1 over the budget range that resists outlier seeds. Cross-dataset comparison uses the Schulze method~\cite{Schulze2011}: for each pair of strategies, we count on how many of the 4 datasets one achieves a higher AUC, and resolve all pairwise outcomes into a complete ranking. The W-T-L column in Table~\ref{tab:results} reports wins, ties, and losses across the 5 head-to-head matchups. We compare six strategies: Random (uniform sampling), Confidence (uncertainty-only, using the combined signal of Eq.~\ref{eq:combined_uncertainty}), MaxHerding~\cite{Bae2024GeneralizedCF} (coverage-only), UHerding~\cite{bae2025uncertainty} (coverage and uncertainty, adapted as in Section~\ref{sec:uherding}), RankFusion (dual-manifold UHerding, Section~\ref{sec:pipeline_aware}), and CAPA (calibrated pipeline-aware, Section~\ref{sec:pipeline_aware}). These baselines span the two main axes of AL: coverage (MaxHerding) and uncertainty (Confidence). MaxHerding generalizes CoreSet and outperforms it on low-budget benchmarks~\cite{Bae2024GeneralizedCF}, making it the stronger coverage representative. Ensemble and MC-dropout acquisition were not evaluated: Bae et al.~\cite{bae2025uncertainty} showed limited impact of the uncertainty estimator on UHerding, and retraining both stages from scratch at each round makes the additional cost prohibitive.

  \subsection{Results}
  \label{sec:results}

  \begin{table}[t]
    \centering
    \caption{Normalized AUC (\%) and per-dataset ranks (in parentheses, 1\,=\,best) computed from median learning curves over 10 seeds. The rightmost columns report the Schulze ranking~\cite{Schulze2011} and each strategy's pairwise record (wins--ties--losses against the 5 other strategies), both derived from cross-dataset AUC comparisons. Best AUC per column in bold.}
    \label{tab:results}
    \small
    \setlength{\tabcolsep}{3pt}
    \begin{tabular}{@{}l rrrr r !{\vrule width 1.2pt} c l@{}}
      \toprule
      Strategy & inv-int & bus-int & fintab & pub1m & Mean & \textbf{Schulze} & \textbf{W-T-L} \\
      & \multicolumn{4}{c}{\scriptsize AUC\,\%\,$\uparrow$ \textit{(rank\,$\downarrow$)}} & \scriptsize$\uparrow$ & \scriptsize$\downarrow$ & \\
      \midrule
      \secrow{Pipeline-aware (ours)} \\
      \quad CAPA       & \auc{64.44}{4} & \auc{54.14}{3} & \aucb{70.52}{1} & \auc{70.86}{3} & 64.99 & \schd{1} & \wtl{1}{3--2--0} \\
      \quad RankFusion & \auc{65.22}{2} & \aucb{57.51}{1} & \auc{70.34}{2} & \auc{69.90}{5} & \textbf{65.74} & \schd{1} & \wtl{1}{3--2--0} \\[3pt]
      \secrow{Adapted (ours)} \\
      \quad UHerding   & \aucb{66.24}{1} & \auc{52.78}{5} & \auc{70.16}{3} & \auc{70.77}{4} & 64.99 & \sch{3} & \wtl{3}{2--2--1} \\[3pt]
      \secrow{Baselines} \\
      \quad Confidence & \auc{64.69}{3} & \auc{52.83}{4} & \auc{68.90}{5} & \aucb{71.99}{1} & 64.60 & \sch{4} & \wtl{4}{1--3--1} \\
      \quad MaxHerding & \auc{63.80}{6} & \auc{56.71}{2} & \auc{69.88}{4} & \auc{69.70}{6} & 65.02 & \sch{5} & \wtl{5}{0--2--3} \\
      \quad Random     & \auc{64.20}{5} & \auc{51.93}{6} & \auc{68.77}{6} & \auc{71.38}{2} & 64.07 & \sch{6} & \wtl{6}{0--1--4} \\
      \bottomrule
      \multicolumn{8}{@{}l}{\scriptsize $\dagger$\,Tied at Schulze rank 1; their head-to-head is 2--2 across datasets.} \\
    \end{tabular}
  \end{table}

\begin{figure}[t]
    \centering
    \includegraphics[width=\linewidth]{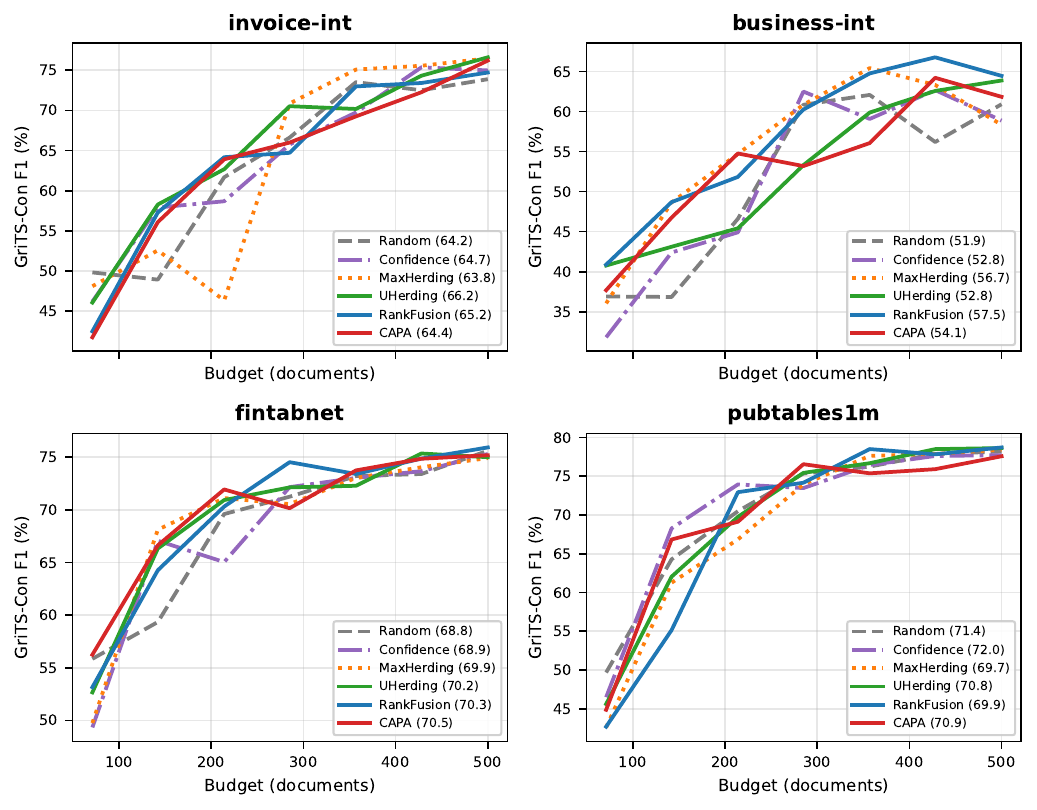}
    \caption{Median learning curves (GriTS-Con F1) across 10 seeds. Dashed/dotted lines denote baselines, solid lines proposed methods. Hybrid methods separate from single-signal baselines early and maintain their advantage throughout the budget range.}
    \label{fig:learning_curves}
  \end{figure}

  Table~\ref{tab:results} and Figure~\ref{fig:learning_curves} report normalized AUC and median learning curves for all strategies. As visible in the learning curves, \textbf{our proposed methods separate from baselines early} (around budget 200) and maintain their advantage throughout the budget range. Since AUC integrates performance over the entire budget trajectory, this early and sustained advantage is directly reflected in the table scores. No single-signal baseline is consistent across datasets: Confidence ranks first on pubtables1m (71.99), where high model performance yields reliable uncertainty estimates, but fifth on fintabnet. MaxHerding ranks second on business-int (56.71), where scarce data makes coverage critical, but sixth on invoice-int. UHerding, our adaptation of coverage-weighted uncertainty sampling to cascaded table extraction, \textbf{ranks above all three baselines in the Schulze cross-dataset ranking} (rank 3).

  Our two pipeline-aware extensions improve further. RankFusion achieves the \textbf{highest mean AUC} (65.74), with its largest gain on business-int (57.51), where extending coverage to both document-level and table-level embedding spaces amplifies the diversity signal. CAPA leads on fintabnet (70.52) and beats UHerding on 3 of 4 datasets, making it the \textbf{most consistent} improvement over the adapted baseline. \textbf{Both extensions tie at Schulze rank 1} (3 wins, 2 ties, 0 losses each), confirming that pipeline-aware sampling yields consistent improvements. Despite their shared rank, they exhibit different risk profiles: RankFusion's coverage-oriented approach is less suited to pubtables1m, where high model performance makes uncertainty the more informative signal. CAPA is never worse than rank 4, making it the safest choice when the difficulty of the target dataset is unknown.

  \section{Discussion}
  \label{sec:discussion}

  \subsection{Datasets Difficulty}
  \label{sec:datasets_difficulty}

  Table~\ref{tab:bottleneck} shows that \textbf{the performance bottleneck is dataset-dependent rather than strategy-dependent}. On invoice-int (TD-bottlenecked), \textbf{UHerding leads}: its uncertainty signal directly targets detection failures, while pipeline-aware methods dilute it across both manifolds. On business-int (balanced, the hardest dataset), \textbf{RankFusion benefits} from dual-manifold coverage. On pubtables1m (TSR-bottlenecked), high absolute performance makes uncertainty reliable, and \textbf{Confidence dominates}. On fintabnet (TSR-bottlenecked), \textbf{CAPA ranks first} by reallocating coverage toward the TSR manifold. CAPA's $\beta$-weighting does not fully compensate on invoice-int: when TD is the true bottleneck, dual-manifold coverage biases selection toward TSR enough to offset per-task calibration.

  \begin{table}[t]
    \centering
    \caption{Per-task bottleneck profiles across datasets. The gap between TD and TSR performance is consistent across strategies and budgets.}
    \label{tab:bottleneck}
    \setlength{\tabcolsep}{6pt}
    \begin{tabular}{lccc}
      \toprule
      Dataset & Bottleneck & TD AP range & TSR AP range \\
      \midrule
      invoice-int   & TD            & 0.54--0.73 & 0.74--0.77 \\
      business-int  & Balanced      & 0.61--0.66 & 0.65--0.68 \\
      fintabnet     & TSR           & 0.85--0.89 & 0.83--0.85 \\
      pubtables1m   & TSR           & 0.80--0.87 & 0.75--0.78 \\
      \bottomrule
    \end{tabular}
  \end{table}

  \begin{figure}[h]
    \centering
    \begin{minipage}[b]{0.48\linewidth}
      \centering
      \begin{tikzpicture}
        \node[anchor=south west, inner sep=0] (img) at (0,0)
          {\includegraphics[width=\linewidth]{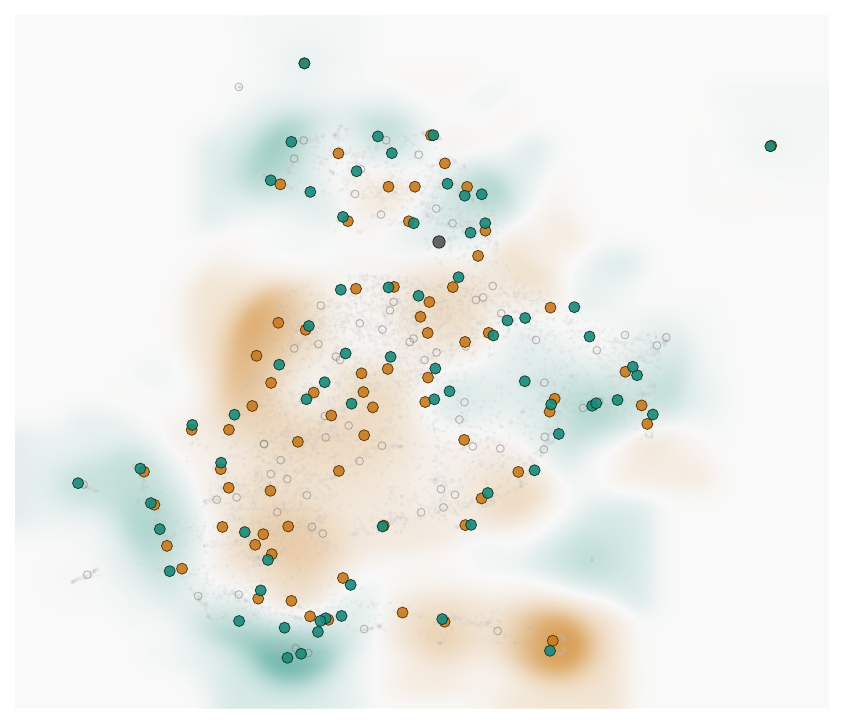}};
        \node[anchor=north west, draw=black!25, fill=white, fill opacity=0.88,
              text opacity=1, rounded corners=1pt, inner sep=2.5pt, font=\tiny]
          at ([shift={(5pt,-5pt)}]img.north west) {%
            \begin{tabular}{@{}l@{}}
              \textcolor[RGB]{26,138,122}{$\bullet$}\,CAPA\\[-1pt]
              \textcolor[RGB]{201,122,31}{$\bullet$}\,MH\\[-1pt]
              $\circ$\,Init
            \end{tabular}};
      \end{tikzpicture}
      \\[-2pt]{\small (a) DiT embedding space}
    \end{minipage}\hfill
    \begin{minipage}[b]{0.48\linewidth}
      \centering
      \begin{tikzpicture}
        \node[anchor=south west, inner sep=0] (img) at (0,0)
          {\includegraphics[width=\linewidth]{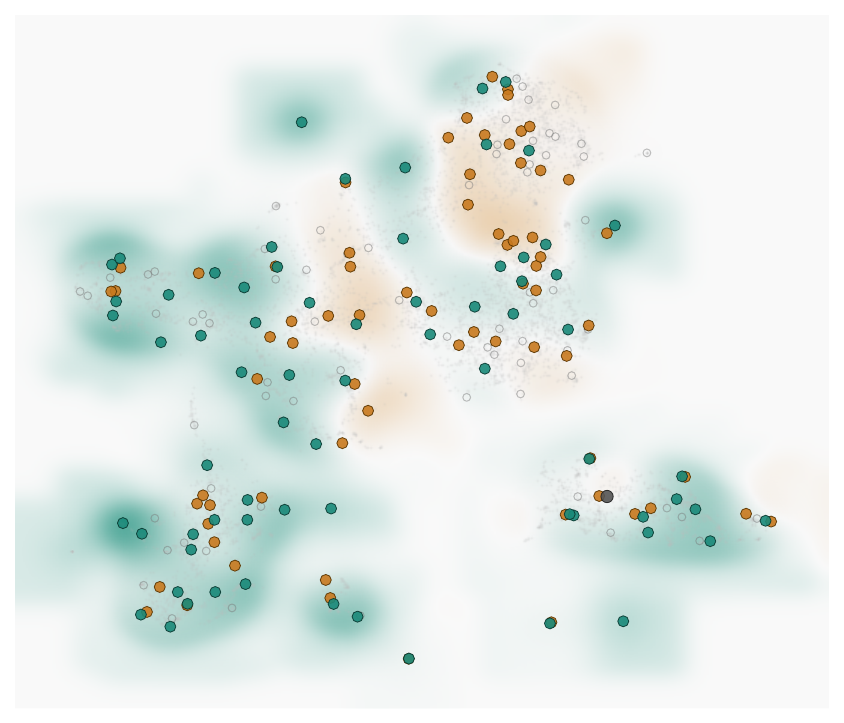}};
        \node[anchor=north west, draw=black!25, fill=white, fill opacity=0.88,
              text opacity=1, rounded corners=1pt, inner sep=2.5pt, font=\tiny]
          at ([shift={(5pt,-5pt)}]img.north west) {%
            \begin{tabular}{@{}l@{}}
              \textcolor[RGB]{26,138,122}{$\bullet$}\,CAPA\\[-1pt]
              \textcolor[RGB]{201,122,31}{$\bullet$}\,MH\\[-1pt]
              $\circ$\,Init
            \end{tabular}};
      \end{tikzpicture}
      \\[-2pt]{\small (b) TSR embedding space}
    \end{minipage}
    \caption{Qualitative comparison of CAPA vs.\ MaxHerding selections on fintabnet (budget 142, first round). Filled dots: \textcolor[RGB]{26,138,122}{green}=CAPA, \textcolor[RGB]{201,122,31}{orange}=MaxHerding, grey=both. Hollow circles: initial set. Background shading: per-pool coverage differential between the two strategies (a:~DiT, b:~TSR).}
    \label{fig:coverage_qualitative}
  \end{figure}

  This reallocation is directly visible in Figure~\ref{fig:coverage_qualitative}. Starting from the same initial labeled set ($|L_0|=71$), CAPA and MaxHerding each select 71 additional samples with near-zero overlap (1 of 71 shared). In the DiT embedding space~(a), both strategies achieve comparable coverage ($\Delta=-0.3\%$, where $\Delta$ is the mean Gaussian kernel coverage gain of CAPA over MaxHerding across the pool). In the TSR embedding space~(b), CAPA covers substantially more of the structural manifold ($\Delta=+3.1\%$). CAPA redistributes its coverage budget toward the space corresponding to the performance bottleneck, which the DiT-only MaxHerding baseline cannot access.

  \subsection{Compute}
  \label{sec:compute}

  All experiments were conducted on a shared GPU cluster using NVIDIA A40, A6000, and H100 GPUs. Each AL run comprises 7 budget levels with 6 sampling rounds, retraining both $M_\text{td}$ and $M_\text{tsr}$ on the augmented labeled set at each budget. A complete run (one strategy, one dataset, all budgets) takes approximately 24 hours. With 6 strategies, 4 datasets, and 10 seeds, the full experimental campaign amounts to 240 runs, totaling approximately 5{,}760 GPU-hours.

  \subsection{Limitations}
  \label{sec:limitations}

  Although evaluating on a single detection architecture is common practice in the AL literature, it does not capture how strategies interact with different model families, particularly YOLO-based detectors whose training dynamics differ significantly from DETR-based models like D-FINE. The experimental scope is restricted to single-table documents without spanning cells, though even under these constraints end-to-end performance remains far from saturated at the budgets considered. The sampling framework operates on page- and table-level embeddings and per-stage confidence rather than on table topology, so it applies unchanged to multi-table pages and spanning-cell layouts. Calibration quality is the open question: multiple tables per page dilute per-document TSR uncertainty, and spanning cells break the regular-grid assumption, both potentially weakening the gating and $\tau$ signals. Whether bottleneck-driven preferences persist under these relaxed constraints remains to be tested. The $\beta$ weights depend on coarse per-stage AP estimates, making them robust to validation noise at early budgets. A formal sensitivity analysis is left to future work. Finally, our budget range (71 to 500 documents) gives dense coverage with 7 points but does not reach high-data regimes where strategy differences may diminish or shift. Scaling further was prohibitive given the cost of the full pipeline.

  \section{Conclusion}
  \label{sec:conclusion}

  We adapted UHerding from image classification to cascaded table extraction with two pipeline-aware extensions: RankFusion, which fuses coverage across document-level and table-level spaces, and CAPA, which adds per-task calibration and adaptive bottleneck weighting. Across 4 datasets and 10 seeds, both share rank 1 in the Schulze ordering, with CAPA offering the most consistent gains. A key finding is that the bottleneck between pipeline stages shapes strategy preferences more than the acquisition function itself: where the pipeline fails matters as much as how samples are selected, suggesting that this framework generalizes to any cascaded detection task with sequential dependencies.

\begin{credits}
\subsubsection{\ackname}
This research was supported by the CIFRE PhD program funded by the ANRT, by the ANR agency under grant number ANR-21-PRRD-0010-01 (LabCom Ideas), as well as by the YOOZ company. This work also benefited from the computing resources of the L3i laboratory, La Rochelle University.

\subsubsection{\discintname}
The authors have no competing interests to declare that are relevant to the content of this article.
\end{credits}

\bibliographystyle{splncs04}
\bibliography{refs}


\end{document}